\documentclass[conference]{IEEEtran}
\IEEEoverridecommandlockouts
\usepackage{cite}
\usepackage{amsmath,amssymb,amsfonts}
\usepackage{algorithmic}
\usepackage{graphicx}
\usepackage{textcomp}
\usepackage{xcolor}
\def\BibTeX{{\rm B\kern-.05em{\sc i\kern-.025em b}\kern-.08em
    T\kern-.1667em\lower.7ex\hbox{E}\kern-.125emX}}
\begin{document}

\title{Classifying Deepfakes Using Swin Transformers\\
}

% \author{Anonymous Authors}

\author{\IEEEauthorblockN{Aprille J. Xi}
\IEEEauthorblockA{\textit{School of Computer Science} \\
\textit{Carnegie Mellon University}\\
Pittsburgh, United States \\
ajxi@andrew.cmu.edu}
\and
\IEEEauthorblockN{Eason Chen}
\IEEEauthorblockA{\textit{School of Computer Science} \\
\textit{Carnegie Mellon University}\\
Pittsburgh, United States \\
EasonC13@cmu.edu}}

% \and
% \and
% \and
% \and

\maketitle

\begin{abstract}
% In an era where digital media authenticity is increasingly critical, the proliferation of deepfake technology poses a serious threat to trust and reliability in online content. This paper explores the use of Swin Transformers—a state-of-the-art architecture based on shifted windows—and Swin hybrid models for the task of deepfake detection and classification. Leveraging the hierarchical and self-attention mechanisms of Swin Transformers, our model aims to efficiently capture image features, enhancing the ability to distinguish subtle manipulations characteristic of deepfakes. We evaluated the model using the Real and Fake Face Detection dataset by Yonsei University's Computational Intelligence Photography Lab, comparing its performance to similar previously tested classifiers. 

The proliferation of deepfake technology poses significant challenges to the authenticity and trustworthiness of digital media, necessitating the development of robust detection methods. This study explores the application of Swin Transformers, a state-of-the-art architecture leveraging shifted windows for self-attention, in detecting and classifying deepfake images. Using the Real and Fake Face Detection dataset by Yonsei University's Computational Intelligence Photography Lab, we evaluate the Swin Transformer and hybrid models such as Swin-ResNet and Swin-KNN, focusing on their ability to identify subtle manipulation artifacts. Our results demonstrate that the Swin Transformer outperforms conventional CNN-based architectures, including VGG16, ResNet18, and AlexNet, achieving a test accuracy of 71.29\%. Additionally, we present insights into hybrid model design, highlighting the complementary strengths of transformer and CNN-based approaches in deepfake detection. This study underscores the potential of transformer-based architectures for improving accuracy and generalizability in image-based manipulation detection, paving the way for more effective countermeasures against deepfake threats.
\end{abstract}


\section{Introduction}
In recent years, the rapid evolution of artificial intelligence has enabled the creation of hyperrealistic manipulated media, commonly referred to as "deepfakes." These synthetic videos, which replace the person's likeness with another using machine learning techniques, pose serious challenges to the integrity and public trust of digital media \cite{toews2020deepfakes}. As such, reliable methods for detecting fake media have become a crucial area of research.

This project aims to tackle the problem of deepfake detection using the Real and Fake Face Detection dataset by Yonsei University's Computational Intelligence Photography Lab, a large collection of images specifically curated to aid in developing deepfake detection models. By applying a state-of-the-art Swin Transformer model \cite{liu2021swin}, as well as exploring novel hybrid approaches such as Swin-KNN and Swin-Resnet, we hope to capture fine-grained spatial and temporal patterns that distinguish real from manipulated facial expressions and movements.

\section{Related Work}
The rapid advancement of deepfake technology has led to research into reliable detection methods. Recent studies highlight deep learning models as effective tools for distinguishing between real and fake media. For example, Qurat et al. used the Real and Fake Face Detection data set with a VGG-16 model, achieving 91.97\% training and 64.49\% test accuracies \cite{nida2021forged}, while Rafique et al. used a Resnet18 and KNN hybrid architecture, achieving 89. 5\% precision \cite{rafique2023deep}. Both models focus on image-based deepfake detection, demonstrating the potential of CNN-based approaches.

Building on these studies, recent work has shifted towards transformer-based architectures, which excel in capturing spatial dependencies in images. Khalid et al., for example, used an encoder-decoder model with a Swin Transformer encoder, achieving an AUC of 0.99 in Celeb-DF and FaceForensics++ videos \cite{khalid2023swynt}. 

Our work aims to improve these methods by applying Swin Transformers and hybrid architectures specifically for deep-fake classification on images. By focusing on the Real and Fake Face Detection dataset, we hope to demonstrate the Swin Transformer's potential for high-accuracy detection in image-based deepfake analysis.

\section{Methods}

This study focuses on classifying deep-fake images by comparing the performance of Swin Transformer-based models against prior CNN-based architectures. All models in this study were trained and tested to ensure consistency, using pre-trained and modified versions to provide a fair and accurate comparison. In line with previous studies, this work incorporates Error Level Analysis (ELA) preprocessing, a technique commonly used in deepfake detection.

ELA preprocessing identifies inconsistencies in JPEG compression, which are often introduced during image manipulations. The method involves re-saving an image at a known compression level and computing the difference between the original and recompressed versions. Mathematically, the difference in ELA can be expressed as 

\begin{equation}
\text{ELA(x)} = | x - \text{JPEG}_{\text{recompressed}}(x)|
\end{equation}

Preprocessed ELA images were then split into test and train datasets, as with prior studies.

The prior CNN-based models were adapted to binary classification tasks with modifications to their architectures and consistent hyperparameter settings. VGG16 was implemented following the research of Qurat et al., using a pre-trained version from Torchvision where the final two layers were replaced with custom CNN layers designed for binary classification. As such, this work also used RMSprop as the optimizer, a learning rate of 0.0001, and a batch size of 5. Batch normalization and dropout were incorporated for regularization to mitigate overfitting. 

While the standalone AlexNet (as well as Resnet18) used a pre-trained model, the hybrid model based on Rafique et al.'s research employed AlexNet as a feature extractor by removing its final classifier layer and replacing it with a custom binary classifier. Dropout was included for regularization to improve robustness. The extracted features were then classified using KNN. Various distance metrics (['cosine', 'euclidean', 'manhattan', 'minkowski', 'chebyshev']) and weight schemes (['uniform', 'distance']) were tested to determine the optimal combination, using RMSprop with a learning rate of 0.0001 and batch size of 4.

The Swin Transformer \cite{liu2021swin} was selected for its innovative architecture, which uses shifted windows in self-attention. Unlike traditional transformers that compute global attention across the entire input, Swin computes attention locally within non-overlapping windows, which shift between layers to enable cross-region interaction. Over multiple layers, this design aggregates local information into a global context, balancing computational efficiency and the ability to capture hierarchical relationships. For this study, the pre-trained Swin Transformer (tiny version) was adapted by removing its classification head and adding a custom binary classifier. Layer normalization and dropout were added for regularization to reduce overfitting. The Swin Transformer was trained with consistent hyperparameters across experiments: a batch size of 32, learning rate of 0.0001, and weight decay of 0.01, ensuring both computational efficiency and fairness in comparisons.

To explore hybrid architectures, this study implemented a Swin and ResNet18 hybrid model, which integrates Swin’s ability to capture global dependencies with ResNet18’s strength in refining local features through residual connections. Swin’s shifted window mechanism captures the broader context of ELA-preprocessed images, while ResNet18 enhances fine-grained detail extraction. Initially, feature fusion was achieved through simple concatenation of features from Swin and ResNet18, but this was later replaced with cross-attention to better integrate interdependent global and local features. Cross-attention enables the model to focus on key interactions between features, improving the hybrid’s ability to detect subtle manipulation artifacts. Like the standalone Swin Transformer, the hybrid model was trained with a batch size of 32, learning rate of 0.0001, and weight decay of 0.01 to ensure consistency.

To address overfitting in the Swin and ResNet18 hybrid, this work attempted freezing all but the last two layers as well as gradually unfreezing layers during training, aiming to stabilize lower-layer features while fine-tuning the top layers. However, this resulted in training stagnation, suggesting that simultaneous fine-tuning of both components is necessary for effective feature fusion. This work also experimented with using Swin’s features in conjunction with KNN for classification, inspired by Rafique et al.’s AlexNet and KNN implementation. This approach heavily overfitted, potentially highlighting Swin’s limitations as a feature extractor.

The motivation for this work was to push the boundaries of the Swin Transformer in deepfake classification and explore how hybrid architectures can enhance detection performance. The Swin and ResNet18 hybrid model, as a novel contribution, demonstrates the potential of combining transformers and CNNs. By comparing these models with established benchmarks and incorporating ELA preprocessing, this study seeks to advance the understanding of transformer-based approaches and their practical applications in manipulation detection.

\section{Results}

We evaluate these models based on Cross-Entropy loss and an accuracy metric defined as follows:

\begin{equation}
\resizebox{0.9\hsize}{!}{$
\text{Accuracy} = \frac{\text{Number of Correct Predictions}}{\text{Total Number of Predictions}} = \frac{TP + TN}{TP + TN + FP + FN}
$}
\end{equation}

Both evaluation metrics were chosen for their normality in classification tasks. Our work and the replicated prior models are summarized in Table 1 and Figure 1. 

\begin{table}[htbp]
\caption{Performance metrics of train and test accuracy across various models and hybrid architectures.}
\begin{center}
\begin{tabular}{|c|c|c|}
\hline
\textbf{Model} & \textbf{Best Train Accuracy}& \textbf{Best Test Accuracy}\\
\hline
Swin & 0.9993 & 0.7129 \\
\hline
Cross-Attention Res-Swin & 0.9958 & 0.7031 \\
\hline
Res-Swin & 1.0000 & 0.6656 \\
\hline
VGG16 & 0.9426 & 0.6493 \\
\hline
AlexNet-KNN & 0.9594 & 0.6362 \\
\hline
ResNet18 & 0.9937 & 0.6248 \\
\hline
AlexNet & 0.9601 & 0.6215 \\
\hline
Swin-KNN & 1.0000 & 0.3295 \\
\hline
\end{tabular}
\label{tab1}
\end{center}
\end{table}

Based on Table \ref{tab1}, previous models such as VGG16 (64. 93\% test accuracy), ResNet18 (62. 48\%), and AlexNet (62. 15\%) showed solid but inferior performance compared to Swin-based models. Our implementation of the AlexNet-KNN hybrid improved slightly over the standalone AlexNet, reflecting gains from combining feature extraction with KNN..

Even still, the fine-tuned Swin Transformer and its hybrids outperformed the prior models, validating the hypothesis that transformer-based architectures excel in capturing subtle manipulation artifacts. The standalone Swin Transformer achieved the highest test accuracy (71.29\%), benefiting from ResNet's feature refinement and Swin's attention mechanisms. It outperformed the hybrid model without cross-attention, perhaps due to cross-attention allowing for better feature integration, improving the model's ability to capture subtle interdependencies. However, Swin-KNN was significantly overfitted, achieving perfect training accuracy but poor test accuracy (32.95\%), highlighting the potential limitations of Swin as a feature extractor for KNN.

\begin{figure*}[htbp]
\centerline{\includegraphics[width=12.5 cm]{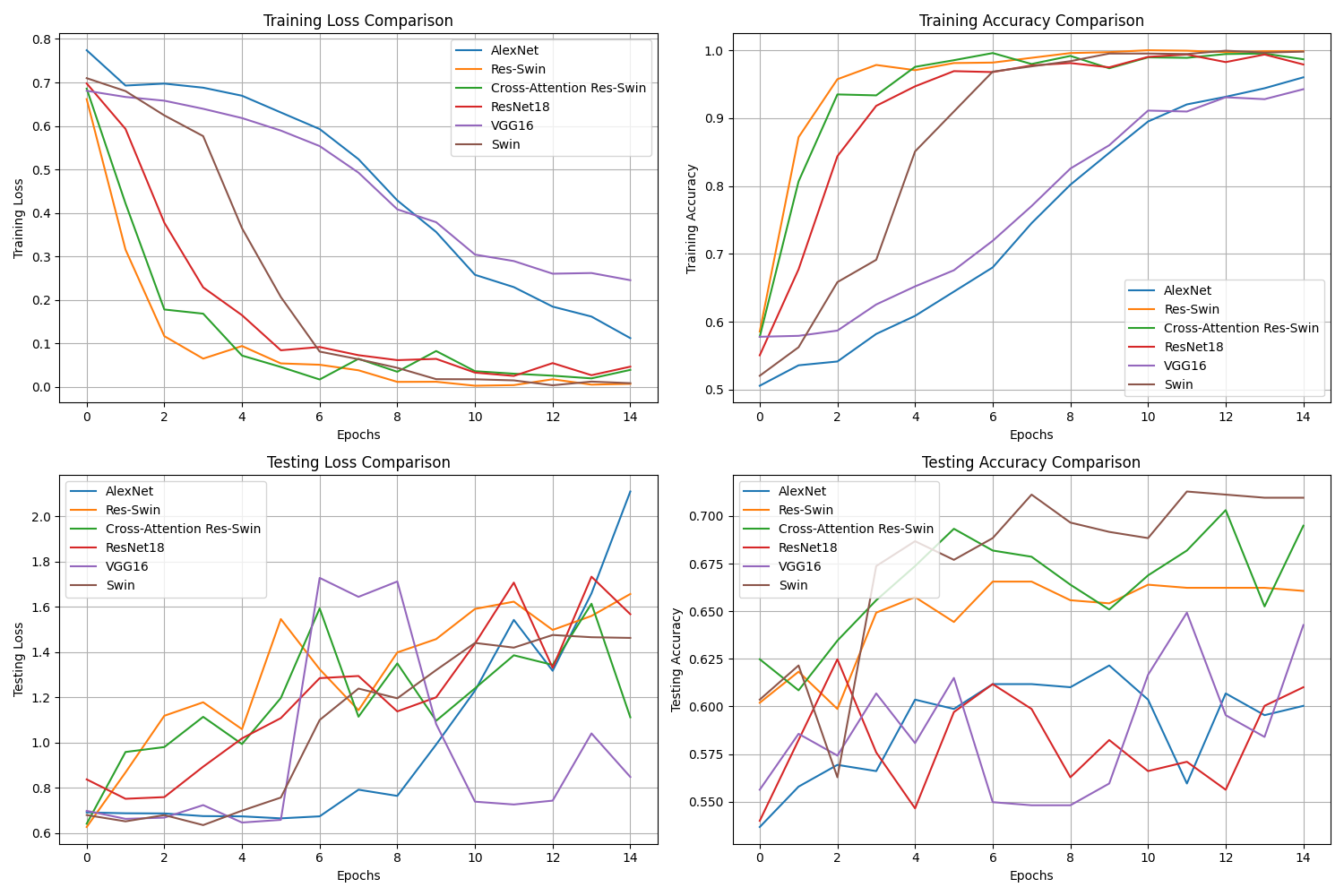}}
\caption{ Training and testing performance comparison of various deepfake classification models, including AlexNet, ResNet18, VGG16, Swin, and hybrid architectures. The top row shows training loss and accuracy, while the bottom row shows testing loss and accuracy.}
\label{fig}
\end{figure*}

Figure \ref{fig} also reveals that while Swin-based models exhibit faster convergence after initial fluctuations, models like AlexNet and VGG16 show more erratic trends, indicating challenges in adapting to the nuanced features of deepfake classification. ResNet18, although solid in training performance, suffers from unstable testing loss, further emphasizing the superior generalization capabilities of Swin-based architectures.

\section{Discussion, Limitation, and Future Works}
The results highlight the effectiveness of the Swin Transformer and its hybrid architectures, leveraging hierarchical attention and feature fusion for deepfake detection. However, limitations remain: the Swin-KNN hybrid overfitted significantly, suggesting the need for better feature extractors, and Swin's computational demands may hinder scalability compared to the efficiency of Rafique et al.'s AlexNet-based approach. The Swin Transformer likely outperformed the Res-Swin models because its fully transformer-based architecture captures global and local dependencies more effectively, avoiding potential loss of information during feature fusion. Still, the Res-Swin model outperformed prior CNN architectures.

A strong assumption was made when selecting ResNet as CNN for hybridization with Swin. This may overlook the potential benefits of other CNNs, such as VGG16, which, despite being computationally heavier, can extract hierarchical features with simplicity, or AlexNet, which is lightweight and faster to train, albeit less expressive for complex tasks. Future work could explore alternative CNNs to determine if ResNet is the most effective.

Another limitation was the use of basic hyperparameters and optimizers during fine-tuning, which may have affected comparison accuracy. This constraint limited the exploration of potential performance gains through more advanced or customized hyperparameter tuning, which could lead to further optimization of model performance. Additionally, although our fine-tuned implementation of VGG16 achieved comparable or slightly better accuracy than the implementation of Qurat et al., our Alexnet + KNN implementation fell short of the reported accuracy of Rafique et al. However, without detailed information on Rafique et al.'s training setup, including optimizers, preprocessing, or other fine-tuning techniques, it is challenging to pinpoint the source of this discrepancy.

Future improvements include applying better regularization techniques to reduce overfitting, experimenting with improved methods for fusing Swin and CNN architectures, and exploring alternative hybrid models. Expanding beyond ELA preprocessing could further enhance robustness across diverse manipulation techniques. Finally, incorporating more sophisticated hyperparameter optimization and benchmarking against a wider range of existing implementations could refine comparisons and advance the practical application of Swin-based models for deepfake detection.

\bibliographystyle{IEEEtran}
\bibliography{reference.bib}

\end{document}